%% file: paper.tex
\documentclass{article}
\usepackage[T1]{fontenc}
\usepackage{spconf,amsmath,graphicx,booktabs}
\usepackage[table]{xcolor}
\definecolor{fgrow}{HTML}{E6F0FC}

\newcommand{\ci}[2]{{\small$[#1,$\allowbreak$\,#2]$}}

\title{KNOWING WHEN TO TRUST A PRIOR:\\ RELIABILITY-GATED CUE FUSION FOR VIDEO GAZE PREDICTION}

\name{Lichen Zhu, Yueqian Lin, Yiheng Wang, Hai ``Helen'' Li, Yiran Chen\thanks{Preprint. Submitted to ICASSP 2027. Supplementary material follows the references.}}
\address{Duke University, Durham, NC, USA}

\begin{document}
\maketitle

\begin{abstract}
Video gaze prediction is led by gaze-trained models, yet gaze-free priors carry signal those models have not absorbed, if one knows when to trust them. We propose FocusGate, a gated ensemble of gaze-free priors whose members may abstain. A per-frame gate reads three shape statistics of a defocus map and selects the frames on which the estimator is above chance on average, so rejected frames reduce to the base exactly, while midrank normalisation lets an all-zero prior abstain at zero parameters. Gated fusion is significantly positive on film, sports and web video, whereas unconditional fusion is harmful on sports and null on web. Added to four supervised predictors, the NTIRE 2026 champion among them, FocusGate improves all sixteen model--domain cells in shuffled AUC, fifteen significantly, one domain pre-registered and scored once, while adding only 1\% to the champion's latency. Alone, it surpasses TASED-Net and UNISAL in shuffled AUC on film with a 16-frame causal mean.
\end{abstract}


\section{Introduction}
\label{sec:intro}

Filmmakers keep the narratively relevant region in focus, and eye-tracking studies confirm that blur contrast steers fixations \cite{smith2012attentional,zhang2019dof}. Modern video saliency is dominated by trained networks
\cite{min2019tased,droste2020unisal,moradi2024salfom,xiong2024diffsal,moskalenko2024aim,moskalenko2026ntire}, yet the explicit priors
used to compose and interpret such systems have hardly changed: a centre Gaussian, faces \cite{vo2012eyes}, motion
\cite{mital2011clustering}. Focusness has served salient-object detection in static images \cite{zhuo2011defocus}, and blur identification has been
injected into saliency models before \cite{baveye2012blur}, but always unconditionally. Reliability-aware fusion has precedents \cite{alshawi2018uncertainty,cong2016depthconf}, but whether such priors still add anything beside gaze-trained models, and on which frames, has not been asked.

A defocus map is a region cue whose reliability varies from frame to frame: it marks the subject in a shallow depth-of-field shot but only texture noise in a deep-focus one. We find that, fused unconditionally, the map of a scratch-trained estimator lowers sAUC on every base we tested (Sec.~\ref{ssec:gatehelp}). In this work we therefore treat trust as a per-frame decision. A reliability gate reads three shape statistics of the map and selects the frames on which the estimator is, on average, above chance. This removes the harm and, with a stronger estimator, yields a significant gain in every domain tested, while on rejected frames the prediction equals its base by construction, i.e., selective prediction applied to cue fusion.

\begin{figure*}[t]
\centering
\includegraphics[width=\textwidth]{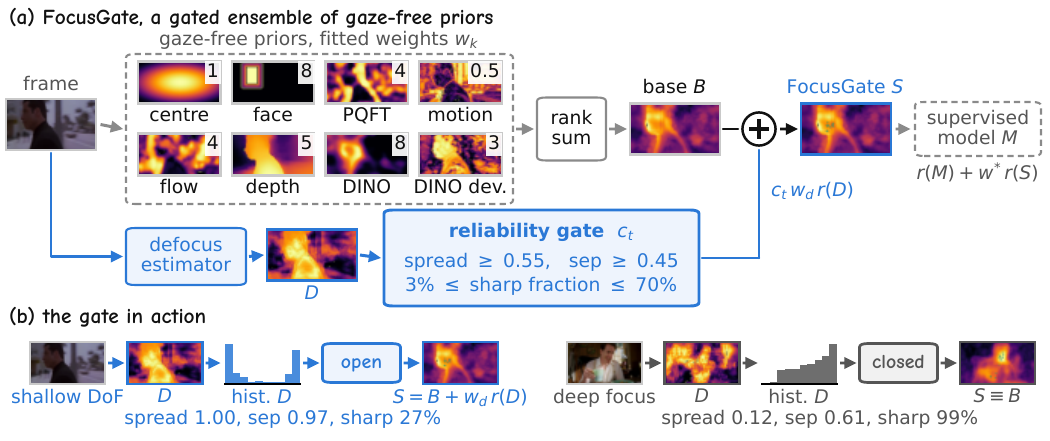}
\vspace{-4mm}
\caption{FocusGate. (a) Eight gaze-free priors (fitted weights $w_k$) are rank-summed into the base $B$. The defocus map $D$ enters only through the gate $c_t$, which thresholds three shape statistics of $D$ and never sees the fixations.  (b) The gate opens on a bimodal $D$ and closes on a unimodal one, where the prediction reduces to $B$ exactly.}
\label{fig:pipe}
\vspace{-3mm}
\end{figure*}

Across seven cue-estimator pairs under one pipeline, we find that gating helps only where an estimator has an identifiable regime in which it collapses to chance, and not merely where it is weak. Two predictions follow, and both are confirmed. A stronger estimator of the same cue \cite{cun2020defocus,jin2023ddffnet,liu2023evp} shrinks the regime and turns the removed harm into a significant gain, and domain shift re-opens it, so that on sports the ungated gain flips to a significant harm while the gate stays positive (Sec.~\ref{ssec:transfer}). We call the resulting system FocusGate, a gated ensemble of gaze-free priors (Fig.~\ref{fig:pipe}) in which each prior may abstain, either through its own output or through the gate.

Our contributions are as follows.
\begin{itemize}\setlength{\itemsep}{0pt}
\item FocusGate, a gated ensemble of gaze-free priors that improves the shuffled AUC of four supervised models, the NTIRE 2026 champion among them, in all sixteen model--domain cells (fifteen significantly, one domain pre-registered and scored once) while adding 1\% to the champion's latency, and that alone surpasses 2019--2020 supervised models in shuffled AUC on film with a 16-frame causal mean (Table~\ref{tab:money}).
\item A per-frame reliability gate for a defocus cue that equals its base on rejected frames, converts the scratch estimator's harm on film to noise ($-0.94^{\dagger}$ to $+0.04$) and flips frozen sports transfer from $-1.40^{\dagger}$ to $+0.68^{*}$ (Secs.~2.1, 4.2, 4.3), and midrank normalisation that keeps an abstaining prior inert at zero parameters (Sec.~4.1).
\end{itemize}

\begin{table*}[t]
\centering
\caption{sAUC (\%) of every predictor under one protocol on held-out video
(Sec.~3). ms/frame per component on one A100 (batch 1, one forward pass per predicted frame, two for ViSAGE's expert pair). A ``$+$ FocusGate'' row adds 111\,ms to its model. Bold with $^{*}$: paired 95\% cluster CI vs.\ the model's own row excludes zero and survives a Holm correction over the sixteen cells.
$^{\ddagger}$: confirmatory domain, pre-registered and scored once (the ViNet-S cell was added afterwards and is exploratory). These cells measure the ensemble, not the gate, whose pre-specified contrast is null (Sec.~\ref{ssec:gatehelp}).}
\label{tab:money}
\small
\input{gen/table1_body}
\vspace{-3.5mm}

\end{table*}

\section{Method}
\label{sec:method}

FocusGate is sketched in Fig.~\ref{fig:pipe}. Given a frame, gaze-free priors are rank-summed into a base map $B$ (Sec.~2.2). A defocus estimator produces an in-focus probability map $D$, and a gate $c_t$ decides from three shape statistics of $D$ whether $D$ enters the sum (Sec.~2.1). 

\textbf{Two estimators of one cue.} Both predict a per-pixel in-focus probability $P$ and are trained on CUHK blur detection \cite{shi2014discriminative} with degradations matched to DivX-era footage, motion blur counting as out-of-focus \cite{kim2018defocus}. The \emph{scratch} estimator is a
plain U-Net (2.76M parameters) with pixel AUROC $0.926$/$0.921$ on
clean/degraded held-out images. The \emph{recipe} estimator adds the three ingredients of modern
defocus-blur detectors: an ImageNet-pretrained encoder
(MobileNetV3-L), train-time depth
distillation \cite{ranftl2020midas,cun2020defocus,jin2023ddffnet}, and a
high-frequency input channel \cite{liu2023evp}, reaching AUROC $0.958$/$0.953$ (15\,ms, Table~\ref{tab:money}).

\subsection{Shallow-DoF reliability gate}
\label{ssec:gate}

From $P$ we compute spread ($p_{90}{-}p_{10}$), separation of the two modes
(mean of $P{>}0.5$ minus mean of $P{\le}0.5$) and the sharp-area fraction.
A frame is \emph{accepted} ($c_t{=}1$) when $\text{spread}\ge0.55$,
$\text{sep}\ge0.45$ and $3\%\le\text{sharp}\le70\%$. The rule reads only the map's shape, and Sec.~\ref{ssec:transfer} shows that the gain it selects lives in the map's spatial content. Thresholds were fixed once on CUHK, before any gaze data was used,
and never re-tuned.

\subsection{Gated rank-fusion}
\label{ssec:fusion}

Let $r(\cdot)$ denote per-frame rank normalisation to $[0,1]$, tied values sharing their midrank, which makes heterogeneous cues scale-free \cite{bylinskii2018metrics}. Cues are ranked individually and the weighted sum is never ranked again, so for a base $B$ and defocus map $D$,
\begin{equation}
S \;=\; B \;+\; c_t\, w_d\, r(D),\qquad B=\textstyle\sum_k w_k\,r(c_k)^{\gamma_k}.
\label{eq:fusion}
\end{equation}
The reference base sums a centre Gaussian, a face prior \cite{yunet} and spectral residual (SR) saliency \cite{hou2007saliency}. When $c_t{=}0$, Eq.~\eqref{eq:fusion} \emph{is} $B$, so off-gate behaviour (frames with $c_t{=}0$, the rest on-gate) is identical by construction. The full ensemble runs Eq.~\eqref{eq:fusion} over the eight priors of Fig.~\ref{fig:pipe}, where PQFT follows \cite{guo2008pqft} and the ranked DINO attention is blurred with $\sigma{=}3.5$ and re-ranked. The weights $w_k$ are those shown in Fig.~\ref{fig:pipe}, $w_d{=}0.25$, $\gamma_k{=}12$ for face, 3 for DINO and 1 elsewhere, and SR and Depth-Anything-V2 (DAv2) \cite{yang2024dav2} were swept to zero, thirteen scalars fitted by coordinate ascent on train sAUC. A supervised model $M$ enters as $r(M)+w^{*}r(S)$ with train-frozen $w^{*}{=}$ 0.4, 0.3, 0.15 and 0.2 for TASED-Net, UNISAL, ViNet-S and ViSAGE. The temporal variant replaces the DINO, flow and defocus maps by the rank of their causal mean over eleven lags spanning 16 frames, refits the scalars and re-admits DAv2 ($+63$\,ms), while the gate still reads the current frame's $D$.

\section{Experimental setup}
\label{sec:setup}

\textbf{Data.} We use Hollywood-2 with the Actions-in-the-Eye recordings \cite{mathe2015actions}, 6.66M fixations rebuilt from the raw 500\,Hz streams, on the standard 823/884 split, scoring the last frame
of up to two clips per video. Transfer uses UCF-Sports
\cite{mathe2015actions} and DHF1K \cite{wang2018dhf1k}, every parameter
frozen before scoring. DIEM \cite{mital2011clustering} is a confirmatory fourth domain, acquired only after every parameter was frozen, pre-registered (the Hollywood-2 protocol, in a timestamped file of the released repository) and scored once (Table~\ref{tab:money}, $^{\ddagger}$ column).

\textbf{Leakage-free subsets.} TASED-Net, UNISAL and ViNet-S saw these datasets' training splits, so every row is scored on held-out video only, while ViSAGE, trained on NTIRE 2026 mouse-tracking data, runs zero-shot. The four baselines span 2019--2026 and 3.7M--6.9B parameters. The held-out sets are the Hollywood-2 test split (1{,}715 frames, 867 videos after 17 undecodable target frames), the UCF-Sports \emph{testing} directory (993 frames, 47 videos, every third frame up to 40 per video), DHF1K 601--700 (1{,}175 frames, 100 videos, twelve evenly spaced frames per video minus 25 without a fixation at scoring resolution), and DIEM (168 frames, 84). TASED-Net and ViNet-S consume a 32-frame clip, UNISAL a recurrent state warmed over up to 96 frames. Our TASED-Net reproduction on the DHF1K validation split gives sAUC
$71.4$ and AUC-J $91.0$ (published test-set values $71.2$ and $89.5$, full-frame protocol), and under that
same all-frame protocol (60{,}273 frames) frozen single-frame FocusGate scores
$72.9$, $+1.5$ \ci{+0.2}{+2.7} over TASED-Net. Code and every table's script will be released.

\textbf{Metrics.} We report shuffled AUC (sAUC, tables in \%, deltas in points) \cite{bylinskii2018metrics}
with negatives drawn from the fixation \emph{distribution} of other videos,
so a predictor exploiting centre bias alone scores chance (Table~\ref{tab:money}, row 1). Density scores such as NSS track a calibration step rather than the cues (Sec.~\ref{ssec:transfer}), so sAUC is the one metric reported. Scores are per-video means averaged over videos, marginalised with their intervals over twenty negative-sampling seeds, and every interval is a 10{,}000-resample percentile bootstrap clustered on videos.
$^{*}$/$^{\dagger}$ mark 95\% CIs
excluding zero (gain/harm). The pre-specified primary contrast is gated vs.\ base for the scratch estimator on the reference base (null, Sec.~\ref{ssec:gatehelp}), and the paired gated-minus-ungated delta is the gate's direct measure. Later arms were each specified, train-tuned and scored once, with unadjusted intervals.

\section{Results}
\label{sec:results}

\subsection{Comparison with prior work}
\label{ssec:money}

Table~\ref{tab:money} puts every predictor on one protocol and on held-out
video only, against TASED-Net \cite{min2019tased}, UNISAL \cite{droste2020unisal}, ViNet-S \cite{chandra2025vinetpp} and the NTIRE 2026 champion ViSAGE \cite{wang2026visage}. Adding the rank-normalised ensemble to each supervised prediction, with one train-frozen weight per model, improves sAUC in all sixteen model--domain cells, and fifteen gains survive the Holm correction.  ViSAGE $+$ FocusGate reaches $81.9$ on film and $79.4$ on
DIEM, the best numbers in the table, and all four DIEM cells are significant. We read this as complementarity rather than strength, since the ensemble alone trails ViNet-S and ViSAGE on every domain yet still lifts both. The exception, ViSAGE on web, is within noise ($+0.03$ \ci{-0.24}{+0.34}).

Gaze enters FocusGate only through its thirteen fusion scalars. Even so, with a 16-frame causal mean the ensemble surpasses TASED-Net and UNISAL in sAUC on film and on the untouched DIEM domain, and its single-frame form scores $79.3$ on sports against ViSAGE's $80.5$ ($-1.2$ \ci{-2.6}{+0.2}). It runs at $111$\,ms per frame, against $69$\,ms for ViNet-S and $9{,}916$\,ms for ViSAGE's two forward passes of a 6.9B backbone with two decoders, so adding it to the champion costs 1\% in latency. Inside the ensemble the defocus prior adds little once the other seven priors are present, and the gate acts as a downside bound, since ungated harm grows with its weight ($-0.22^{\dagger}$ at $w{=}2$ on sports) while the gated term never leaves noise ($-0.05$ \ci{-0.14}{+0.05}).

\textbf{Abstention.} One prior signals failure without any gate. The face detector returns nothing on $33\%$/$74\%$/$69\%$ of film/sports/web frames, and its prior is then a constant map, i.e., an abstention. Tie-splitting rank normalisation destroys it by fabricating a full-contrast pattern at the cue's weight, whereas shared-midrank ranking keeps a constant map constant, so the cue self-abstains. This costs zero parameters and is worth $+0.6$ points at frozen weights, five times the explicit face gate it replaces ($+0.12$ \ci{+0.04}{+0.21}).

\begin{table}[t]
\centering
\caption{The gate on seven cue-estimator pairs (H2 test split, reference base centre+face+SR, $w{=}0.5$). Shaded: the defocus cue that FocusGate gates, with the recipe estimator also transferred frozen. Unshaded: the same rule on other cues (motion: spread $\ge0.1$, moving fraction $\le0.35$, train-selected). rate: accept rate. on/off: accepted/rejected frames.}
\label{tab:panel}
\small
\setlength{\tabcolsep}{2.6pt}
\input{gen/table2_body}
\vspace{-1.5mm}
\end{table}

\begin{figure}[t]
\centering
\includegraphics[width=\columnwidth]{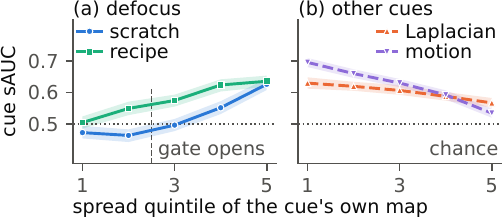}
\vspace{-3mm}
\caption{The collapse regime. Standalone sAUC of each cue on the 1{,}715 test frames by quintile of the spread of its own map (shaded: 95\% CI). Defocus maps sit at chance when unimodal and rise when bimodal, while sharpness and motion never approach chance.}
\label{fig:diag}
\vspace{-4mm}
\end{figure}

\subsection{When does the gate help?}
\label{ssec:gatehelp}

With the scratch estimator, unconditional fusion harms every base composition we tested and the gate removes the harm on all of them. On the reference base (centre+face+SR,
$69.0$) the pair is $-0.94^{\dagger}$ ungated against $+0.04$
\ci{-0.12}{+0.21} gated (Table~\ref{tab:panel}), i.e., a significant loss converted to a null, with a paired gated-minus-ungated contrast of $+0.99$ \ci{+0.74}{+1.24}.

Table~\ref{tab:panel} then applies the same rule, motion excepted (see caption), to seven cue-estimator pairs, and gating does not reward cue strength. A Laplacian sharpness map outpredicts the scratch defocus map standalone ($60.1$ vs.\ $52.2$) yet harms under either policy, and for motion and PQFT, both stronger still, the gate only discards signal. What separates the rows is the off-gate column, and Fig.~\ref{fig:diag} shows why. Binned by the spread of their own map, the defocus estimators fall to chance where the map is unimodal and rise above $60$ where it is bimodal, whereas sharpness and motion never approach chance, so the rule has nothing to select. We conclude that the rule is a defocus-reliability test and the other rows its negative control, where it is inert or harmful.

A better estimator shrinks the collapse regime but does not remove the need for the gate (Table~\ref{tab:panel}, row 3). Its off-gate sAUC rises from $48.0$ to $53.9$, ungated fusion becomes safe ($+0.36^{*}$) and gating still adds ($+0.60^{*}$), within noise on film ($+0.23$ \ci{-0.02}{+0.49}) but decisive under transfer.

\subsection{Frozen transfer and controls}
\label{ssec:transfer}

The two frozen rows of Table~\ref{tab:panel} transfer the recipe estimator and the reference base to sports and web video with every parameter frozen. Domain shift re-opens the collapse regime and the gate finds it. On sports, where the recipe map alone is barely above chance ($50.7$), ungated fusion flips to a significant harm ($-1.40^{\dagger}$) while gated fusion stays significantly positive ($+0.68^{*}$), and on web the ungated term is null ($-0.08$) while the gated one is again positive ($+0.57^{*}$), so the thresholds transfer without re-tuning.

\textbf{Controls.} Five checks on the H2 test split rule out alternatives. A random gate at the same accept rate trails the real gate by $0.35$ to $0.56$ points over twenty draws, so the gain is not an artefact of down-weighting the cue. Permuting the map's pixels on accepted frames turns the gain into harm ($+1.54^{*}$ vs.\ $-1.27^{\dagger}$), so it lives in the map's spatial content rather than in the frame selection. A soft gate performs identically to the hard one ($-0.01$ \ci{-0.05}{+0.04}), and the gain holds in both extreme motion quartiles ($+0.33^{*}$/$+0.89^{*}$) and across fifteen recalibrations of the map ($+0.52$ to $+0.64$, all significant). Density calibration lifts film NSS from $1.81$ to $2.69$ at unchanged AUC-Judd, so the priors carry ordering, not density.

\section{Conclusion}
\label{sec:concl}

We proposed FocusGate, a gated ensemble of gaze-free priors in which each prior may abstain. Its gate reads three shape statistics of the defocus map and selects the frames on which the estimator is above chance on average. This removes the harm of unconditional fusion and, with a stronger estimator, yields a significant gain in every domain tested. Added to four supervised predictors, the NTIRE 2026 champion among them, the ensemble improves every model's sAUC in every domain, a pre-registered one included. The gate is specific to defocus, the only cue tested with a collapse regime, and future work will look for reliability statistics of other cues.

\section{Compliance with Ethical Standards}
\label{sec:ethics}

This retrospective study uses only public eye-tracking data, namely Hollywood-2 with the Actions-in-the-Eye recordings and UCF-Sports \cite{mathe2015actions}, DHF1K \cite{wang2018dhf1k} and DIEM \cite{mital2011clustering} (CC BY-NC-SA 3.0). No new human or animal data was collected and no ethical approval was required.

{\small
\bibliographystyle{IEEEbib}
\bibliography{refs}
}

\input{supp_body}

\end{document}

%% file: gen/table1_body.tex
\begin{tabular*}{\textwidth}{@{\extracolsep{\fill}}llccccc@{}}
\toprule
predictor & year & ms/frame & Film (H2) & Sports (UCF) & Web (DHF1K) & DIEM$^{\ddagger}$ \\
\midrule
centre prior (chance calibration) & --- & $<$0.1 & 49.9 & 51.4 & 50.6 & --- \\
flow residual, camera-comp. & --- & 19.8 & 63.7 & 69.9 & 60.2 & --- \\
face detector \cite{yunet} & 2023 & 12.1 & 65.3 & 52.9 & 53.9 & --- \\
depth, MiDaS \cite{ranftl2020midas} & 2022 & 11.1 & 61.5 & 63.2 & 58.9 & --- \\
DINOv2 attention \cite{oquab2023dinov2,darcet2024registers} & 2024 & 51.5 & 71.0 & 74.6 & 66.9 & --- \\
defocus, recipe estimator & ours & 15.4 & 57.6 & 50.7 & 54.8 & --- \\
\midrule
FocusGate, uniform weights & ours & 174 & 73.8 & 77.0 & 67.9 & 72.2 \\
FocusGate, single frame & ours & 111 & 77.8 & 79.3 & 71.3 & 75.5 \\
FocusGate, $+$causal context & ours & 174 & 80.0 & 78.3 & 72.7 & 76.8 \\
\midrule
TASED-Net \cite{min2019tased} (21M) & 2019 & 50 & 78.4 & 77.0 & 72.6 & 74.0 \\
\quad $+$ FocusGate &  &  & \textbf{79.9}$^{*}$ & \textbf{79.8}$^{*}$ & \textbf{73.6}$^{*}$ & \textbf{76.6}$^{*}$ \\
UNISAL \cite{droste2020unisal} (3.7M) & 2020 & 172 & 77.8 & 80.5 & 69.7 & 73.9 \\
\quad $+$ FocusGate &  &  & \textbf{79.0}$^{*}$ & \textbf{81.3}$^{*}$ & \textbf{71.2}$^{*}$ & \textbf{75.4}$^{*}$ \\
ViNet-S \cite{chandra2025vinetpp} (9.5M) & 2025 & 69 & 80.7 & 81.2 & 73.5 & 77.3 \\
\quad $+$ FocusGate &  &  & \textbf{81.4}$^{*}$ & \textbf{81.8}$^{*}$ & \textbf{74.0}$^{*}$ & \textbf{78.2}$^{*}$ \\
ViSAGE \cite{wang2026visage} (6.9B) & 2026 & 9,916 & 81.3 & 80.5 & 75.8 & 78.4 \\
\quad $+$ FocusGate &  &  & \textbf{81.9}$^{*}$ & \textbf{81.7}$^{*}$ & 75.9 & \textbf{79.4}$^{*}$ \\
\bottomrule
\end{tabular*}

%% file: gen/table2_body.tex
\begin{tabular}{lrcccrr}
\toprule
 & & \multicolumn{3}{c}{cue sAUC (\%)} & \multicolumn{2}{c}{$\Delta$ vs.\ base} \\
\cmidrule(lr){3-5}\cmidrule(l){6-7}
cue, estimator & rate & alone & on & off & ungated & gated \\
\midrule
\rowcolor{fgrow} defocus, classical & 0\% & 45.6 & --- & 45.6 & $-1.88$$^{\dagger}$ & $+0.00$ \\
\rowcolor{fgrow} defocus, scratch & 38\% & 52.2 & 58.5 & 48.0 & $-0.94$$^{\dagger}$ & $+0.04$ \\
\rowcolor{fgrow} defocus, recipe & 40\% & 57.6 & 62.9 & 53.9 & $+0.36$$^{*}$ & $+0.60$$^{*}$ \\
\rowcolor{fgrow} \quad frozen, sports & 38\% & 50.7 & 55.7 & 46.9 & $-1.40$$^{\dagger}$ & $+0.68$$^{*}$ \\
\rowcolor{fgrow} \quad frozen, web & 55\% & 54.8 & 57.5 & 51.4 & $-0.08$ & $+0.57$$^{*}$ \\
\midrule
sharpness, Laplacian & 17\% & 60.1 & 57.7 & 60.4 & $-0.67$$^{\dagger}$ & $-0.18$$^{\dagger}$ \\
motion, frame diff. & 56\% & 62.1 & 59.1 & 65.7 & $+0.68$$^{*}$ & $+0.03$ \\
depth, MiDaS & 100\% & 61.5 & 61.5 & --- & $+1.95$$^{*}$ & $+1.95$$^{*}$ \\
PQFT & 0.3\% & 66.8 & --- & 66.9 & $+0.54$$^{*}$ & $+0.00$ \\
\bottomrule
\end{tabular}

%% file: supp_body.tex
\clearpage
\onecolumn
\section*{Supplementary Material}
Every value below is produced by \texttt{src/gen\_supplement.py} from the result files that also feed Tables~1 and 2, under the protocol of Sec.~3 (per-video means, twenty negative-sampling seeds, 10{,}000-resample percentile bootstraps clustered on videos). Nothing here was tuned; the supplement reports intervals, controls and negative results that the four-page limit left out.
\par\noindent\begin{minipage}{\textwidth}
\subsection*{A. All-frame protocol on DHF1K 601--700}
Sec.~3 quotes one number from this run. Both predictors were scored on every annotated frame of the 100 validation videos (60,273 frames), with the DHF1K benchmark's shuffled-AUC mechanics (negatives from the other 99 videos' fixations) and AUC-Judd at the 640$\times$360 ground-truth resolution. The gate accepted 54.9\% of all frames (55\% under the sparse protocol of Table~2).
\begin{center}\small\begin{tabular}{lccc}\toprule predictor & sAUC (\%) & AUC-J (\%) & paired $\Delta$ sAUC vs.\ TASED-Net \\ \midrule
TASED-Net (sliding 32-frame window) & 71.4 & 91.0 & --- \\
FocusGate, single frame, frozen & 72.9 & 84.0 & $+1.5$ [$+0.2$, $+2.7$]$^{*}$ ($n=100$ videos) \\
\bottomrule\end{tabular}\end{center}
FocusGate ranks fixations better than TASED-Net under this protocol while scoring lower on AUC-Judd, which rewards density and centre mass. The same ordering-versus-density split appears in Sec.~E.
\end{minipage}\par\medskip
\par\noindent\begin{minipage}{\textwidth}
\subsection*{B. The sixteen ``$+$ FocusGate'' cells of Table~1 with intervals}
Paired per-video deltas of model $+$ FocusGate over the model alone, 95\% cluster-bootstrap intervals, the two-sided bootstrap $p$ (floored at $1/B$) and the Holm threshold it is compared against. Fifteen of sixteen cells survive the correction; the ViNet-S DIEM cell was added after the DIEM pre-registration.
\begin{center}\small\begin{tabular}{llrcc}\toprule domain & model & $\Delta$ sAUC [95\% CI] & $p$ & Holm \\ \midrule
Film (H2) & TASED-Net & $+1.59$ [$+1.27$, $+1.92$]$^{*}$ & $0.0001$ & yes ($\le 0.0031$) \\
Film (H2) & UNISAL & $+1.26$ [$+1.04$, $+1.49$]$^{*}$ & $0.0001$ & yes ($\le 0.0036$) \\
Film (H2) & ViNet-S & $+0.64$ [$+0.48$, $+0.79$]$^{*}$ & $0.0001$ & yes ($\le 0.0038$) \\
Film (H2) & ViSAGE & $+0.62$ [$+0.42$, $+0.83$]$^{*}$ & $0.0001$ & yes ($\le 0.0033$) \\
Sports (UCF) & TASED-Net & $+2.77$ [$+2.21$, $+3.33$]$^{*}$ & $0.0001$ & yes ($\le 0.0042$) \\
Sports (UCF) & UNISAL & $+0.84$ [$+0.47$, $+1.23$]$^{*}$ & $0.0001$ & yes ($\le 0.0050$) \\
Sports (UCF) & ViNet-S & $+0.65$ [$+0.45$, $+0.85$]$^{*}$ & $0.0001$ & yes ($\le 0.0056$) \\
Sports (UCF) & ViSAGE & $+1.18$ [$+0.81$, $+1.57$]$^{*}$ & $0.0001$ & yes ($\le 0.0045$) \\
Web (DHF1K) & TASED-Net & $+0.98$ [$+0.49$, $+1.50$]$^{*}$ & $0.0002$ & yes ($\le 0.0125$) \\
Web (DHF1K) & UNISAL & $+1.49$ [$+1.06$, $+1.96$]$^{*}$ & $0.0001$ & yes ($\le 0.0063$) \\
Web (DHF1K) & ViNet-S & $+0.48$ [$+0.26$, $+0.72$]$^{*}$ & $0.0001$ & yes ($\le 0.0071$) \\
Web (DHF1K) & ViSAGE & $+0.03$ [$-0.24$, $+0.34$] & $0.8356$ & no ($\le 0.0500$) \\
DIEM & TASED-Net & $+2.61$ [$+1.45$, $+3.92$]$^{*}$ & $0.0001$ & yes ($\le 0.0083$) \\
DIEM & UNISAL & $+1.55$ [$+0.97$, $+2.21$]$^{*}$ & $0.0001$ & yes ($\le 0.0100$) \\
DIEM & ViNet-S & $+0.92$ [$+0.39$, $+1.47$]$^{*}$ & $0.0008$ & yes ($\le 0.0250$) \\
DIEM & ViSAGE & $+0.92$ [$+0.40$, $+1.49$]$^{*}$ & $0.0004$ & yes ($\le 0.0167$) \\
\bottomrule\end{tabular}\end{center}
\end{minipage}\par\medskip
\par\noindent\begin{minipage}{\textwidth}
\subsection*{C. Table~2 with intervals}
The same rows as Table~2, with the 95\% intervals of the two deltas. Reference base centre+face+SR, fusion weight $w{=}0.5$, H2 test split; the two frozen rows transfer the recipe estimator with every parameter fixed.
\begin{center}\small\begin{tabular}{lrrr}\toprule cue, estimator & rate & ungated $\Delta$ [95\% CI] & gated $\Delta$ [95\% CI] \\ \midrule
defocus, classical & 0\% & $-1.88$ [$-2.16$, $-1.61$]$^{\dagger}$ & $+0.00$ [$+0.00$, $+0.00$] \\
defocus, scratch & 38\% & $-0.94$ [$-1.25$, $-0.64$]$^{\dagger}$ & $+0.04$ [$-0.12$, $+0.21$] \\
defocus, recipe & 40\% & $+0.36$ [$+0.05$, $+0.67$]$^{*}$ & $+0.60$ [$+0.44$, $+0.76$]$^{*}$ \\
\quad frozen, sports & 38\% & $-1.40$ [$-2.70$, $-0.13$]$^{\dagger}$ & $+0.68$ [$+0.07$, $+1.36$]$^{*}$ \\
\quad frozen, web & 55\% & $-0.08$ [$-0.75$, $+0.60$] & $+0.57$ [$+0.18$, $+1.01$]$^{*}$ \\
\midrule
sharpness, Laplacian & 17\% & $-0.67$ [$-0.86$, $-0.48$]$^{\dagger}$ & $-0.18$ [$-0.26$, $-0.11$]$^{\dagger}$ \\
motion, frame diff. & 56\% & $+0.68$ [$+0.45$, $+0.91$]$^{*}$ & $+0.03$ [$-0.13$, $+0.20$] \\
depth, MiDaS & 100\% & $+1.95$ [$+1.69$, $+2.23$]$^{*}$ & $+1.95$ [$+1.69$, $+2.23$]$^{*}$ \\
PQFT & 0.3\% & $+0.54$ [$+0.34$, $+0.74$]$^{*}$ & $+0.00$ [$-0.00$, $+0.01$] \\
\bottomrule\end{tabular}\end{center}
\end{minipage}\par\medskip
\par\noindent\begin{minipage}{\textwidth}
\subsection*{D. Gate controls (H2 test split, recipe estimator unless stated)}
\begin{center}\small\begin{tabular}{lr}\toprule control & value \\ \midrule
paired gated $-$ ungated, recipe & $+0.23$ [$-0.02$, $+0.49$] \\
paired gated $-$ ungated, scratch & $+0.99$ [$+0.74$, $+1.24$]$^{*}$ \\
random gate at the same accept rate (39.7\%), 20 draws, gated $\Delta$ vs.\ base & $+0.04$ to $+0.25$ (real gate $+0.60$) \\
real gate $-$ random gate, mean / min over draws & $+0.45$ / $+0.35$ \\
accepted frames ($n=465$ videos): real map vs.\ base & $+1.54$ [$+1.12$, $+1.96$]$^{*}$ \\
accepted frames: pixel-permuted map vs.\ base & $-1.27$ [$-1.59$, $-0.96$]$^{\dagger}$ \\
accepted frames: real $-$ permuted & $+2.81$ [$+2.33$, $+3.31$]$^{*}$ \\
soft gate (logistic) $-$ hard gate & $-0.01$ [$-0.05$, $+0.04$] \\
gated $\Delta$, lowest / highest motion quartile & $+0.33$ [$+0.09$, $+0.58$]$^{*}$ / $+0.89$ [$+0.41$, $+1.37$]$^{*}$ \\
accept rate by motion quartile (low to high) & 30\% / 36\% / 42\% / 52\% \\
gated $\Delta$ over 15 affine recalibrations $a P + b$ of the map & $+0.52$ to $+0.64$, all lower bounds $>0$ \\
gate decision agreement, adjacent frames / 16 frames apart & 92.8\% / 82.3\% \\
\bottomrule\end{tabular}\end{center}
The accept rate rises with motion, yet the gated gain is significant in both extreme quartiles, so the gate is not a motion detector in disguise. The recalibrations rescale and shift $P$ before the three statistics are computed, with the thresholds left untouched.
\end{minipage}\par\medskip
\par\noindent\begin{minipage}{\textwidth}
\subsection*{E. Density-sensitive metrics on the film test split}
sAUC is the paper's metric because a centre-bias predictor scores chance on it. The table shows what the other common metrics say for the same predictions: FocusGate's raw rank-sum map carries ordering but little density, a single train-fitted monotone calibration of its values lifts NSS, CC and SIM without touching AUC-Judd (a ranking metric), and the supervised models remain far ahead on density.
\begin{center}\small\begin{tabular}{lcccc}\toprule predictor & NSS & CC & SIM & AUC-J (\%) \\ \midrule
FocusGate, raw rank sum & 1.81 & 0.292 & 0.108 & 87.4 \\
FocusGate, density-calibrated & 2.69 & 0.409 & 0.234 & 87.4 \\
TASED-Net & 3.59 & 0.540 & 0.368 & 93.2 \\
UNISAL & 3.51 & 0.523 & 0.358 & 92.5 \\
ViNet-S & 3.87 & 0.590 & 0.390 & 94.2 \\
\bottomrule\end{tabular}\end{center}
\end{minipage}\par\medskip
\par\noindent\begin{minipage}{\textwidth}
\subsection*{F. Variants that did not help (fitted on train, frozen sets scored once)}
Each variant was refitted on Hollywood-2 train with the Sec.~2.2 procedure and, where it passed on train, scored once on the frozen sets against the released ensemble. None was adopted.
\begin{center}\footnotesize\setlength{\tabcolsep}{4pt}\begin{tabular}{lcccc}\toprule variant & train sAUC & film $\Delta$ & sports $\Delta$ & web $\Delta$ \\ \midrule
released ensemble (DINOv2-registers base) & 78.40 & --- & --- & --- \\
DINOv2-registers large as DINO cue & 77.73 & $-0.50$ [$-0.73$, $-0.28$]$^{\dagger}$ & $-0.61$ [$-1.14$, $-0.03$]$^{\dagger}$ & $-0.52$ [$-1.03$, $-0.01$]$^{\dagger}$ \\
flip test-time averaging of DINO cue & 78.43 & $+0.01$ [$-0.02$, $+0.04$] & $-0.06$ [$-0.13$, $-0.00$] & $-0.02$ [$-0.07$, $+0.03$] \\
\bottomrule\end{tabular}\end{center}
Also rejected on train alone: per-cue causal temporal pooling for every cue (gain 0 over the released ensemble), uncertainty-weighted fusion with per-frame entropy (the fitted weight of the uncertainty term is 0), and a DINO-internal failure detector (the best internal statistic predicts the frames on which dropping the DINO term would help with ROC-AUC 0.56 on train, below the 0.6 needed to fit a rule). V-JEPA~2 attention and feature maps were texture-like and were not fitted as a cue.
\end{minipage}\par\medskip
\par\noindent\begin{minipage}{\textwidth}
\subsection*{G. Fitted scalars}
Coordinate ascent on train sAUC over the grids listed in Sec.~H; weights multiply the rank-normalised cue after its exponent; the centre Gaussian has weight 1 and is the reference. Cues with weight 0 were swept and rejected.
\begin{center}\small\begin{tabular}{lcc}\toprule cue & single frame & $+$causal context \\ \midrule
face & 8 & 8 \\
spectral residual & 0 & 0 \\
PQFT & 4 & 3 \\
frame-diff motion & 0.5 & 0 \\
flow residual & 4 & 5 \\
MiDaS depth & 5 & 3 \\
Depth-Anything-V2 & 0 & 0.25 \\
DINOv2-registers attention & 8 & 8 \\
DINO local deviation & 3 & 2 \\
gated defocus ($w_d$) & 0.25 & 1 \\
face exponent / DINO exponent & 12 / 3 & 12 / 4 \\
DINO blur $\sigma$ (rank-map pixels) & 3.5 & 3 \\
causal lags (frames) & --- & 0, 1, 2, 3, 4, 5, 6, 8, 10, 12, 16 \\
train sAUC & 78.40 & 80.41 \\
\bottomrule\end{tabular}\end{center}
Supervised fusion weight $w^{*}$ (one per model, train-frozen): TASED-Net 0.4, UNISAL 0.3, ViNet-S 0.15, ViSAGE 0.2.
\end{minipage}\par\medskip
\par\noindent\begin{minipage}{\textwidth}
\subsection*{H. Implementation details}
Settings as they stand in the released scripts (file names in parentheses). Where no fixed random seed was used, the table says so.
\begin{center}\small\begin{tabular}{p{0.24\textwidth}p{0.69\textwidth}}\toprule item & setting \\ \midrule
\multicolumn{2}{l}{\emph{Defocus estimators} (\texttt{train\_dbd.py}, \texttt{train\_dbd\_v2.py})} \\
data and split & CUHK blur-detection images with ground truth (defocus and motion-blur images, any blur counts as not in focus); file stems sorted, shuffled once with a fixed seed (42), the first fifth held out for validation, the rest for training; the same split for both estimators. No image from any gaze dataset. \\
input & $320{\times}320$ RGB in $[0,1]$; target $1-$mask, white = blurred. \\
degradation (training only, per image) & downscale by a factor in $U(0.25,0.9)$ and upscale back ($p{=}0.7$); JPEG at quality $U(25,90)$ ($p{=}0.7$); contrast $U(0.7,1.3)$ and brightness $U(-20,20)$ ($p{=}0.5$); gamma $U(0.7,1.4)$ ($p{=}0.3$); Gaussian noise with $\sigma \sim U(2,8)$ ($p{=}0.3$). Validation images are undegraded; the ``degraded'' AUROC applies the same recipe with a fixed seed. \\
scratch estimator & U-Net, 24/48/96/192/192 channels, four max-pool levels, 2.76M parameters; loss BCE $+\,0.5\,$Dice; AdamW, lr $3{\times}10^{-4}$, weight decay $10^{-4}$, batch 32, 40 epochs, cosine schedule; checkpoint with the best validation AUROC. \\
recipe estimator & MobileNetV3-Large encoder (ImageNet weights), first convolution inflated to four input channels with the new channel zero-initialised; fourth channel = $3{\times}3$ Laplacian magnitude of the grey image, max-normalised per image; auxiliary head regresses MiDaS-small depth precomputed on CUHK (train time only, L1 weight 0.3); loss boundary-weighted BCE $+\,0.5\,$Dice $+\,0.3\,$depth L1; AdamW, lr $5{\times}10^{-5}$ (encoder) and $3{\times}10^{-4}$ (rest), batch 24, 40 epochs, cosine; 4.64M parameters. \\
seeds & no fixed training seed (single run each); the validation loader and the degraded-validation pass use seed 0. \\
\midrule \multicolumn{2}{l}{\emph{Gate} (\texttt{focusgate/dbd.py})} \\
statistics & on the $96{\times}56$ in-focus probability map $P$: spread $=p_{90}-p_{10}$; sep $=$ mean$(P{>}0.5)-$mean$(P{\le}0.5)$; sharp fraction $=$ share of $P{>}0.5$. \\
thresholds & spread $\ge 0.55$, sep $\ge 0.45$, $0.03 \le$ sharp $\le 0.70$; constants set by hand on CUHK validation maps before any gaze data was used, never changed afterwards; the gated gain is flat over spread $\in\{0.45,\dots,0.65\}$ $\times$ sep $\in\{0.35,\dots,0.55\}$ with the sharp bounds fixed (\texttt{oral\_hardening.py}). \\
\bottomrule\end{tabular}\end{center}
\end{minipage}\par\medskip
\par\noindent\begin{minipage}{\textwidth}
\begin{center}\small\begin{tabular}{p{0.24\textwidth}p{0.69\textwidth}}\toprule item & setting \\ \midrule
\multicolumn{2}{l}{\emph{Cues} (\texttt{diem\_extract\_recs.py} is the reference implementation of the full per-frame recipe)} \\
frame geometry & every frame resized to $384{\times}224$; every cue map and fixation mask at $96{\times}56$ (masks by nearest-neighbour downsampling of the fixation image). \\
centre & anisotropic Gaussian, $\sigma_x = 0.28\,W$, $\sigma_y = 0.24\,H$, hard-coded, not tuned on any gaze split (\texttt{focusgate/priors.py}). \\
face & YuNet via OpenCV \texttt{FaceDetectorYN}, long side scaled to 320 px, score threshold 0.6, NMS 0.3; each box filled with its score on the $96{\times}56$ grid, $5{\times}5$ Gaussian blur; midrank; all-zero map when nothing is detected. \\
spectral residual & Hou and Zhang at 192 px working width (width selected on the train split), $3{\times}3$ box filter on the log spectrum. \\
PQFT & quaternion phase spectrum at $128{\times}64$ with frame $t{-}3$ as the motion channel. \\
motion & grey frames blurred $5{\times}5$; max of the absolute differences $|g_t-g_{t-1}|$, $|g_{t-1}-g_{t-2}|$; max-normalised. \\
flow residual & Farneback flow ($0.5, 3, 21, 3, 5, 1.2$) between $t{-}2$ and $t$ at $384{\times}216$; median flow vector subtracted (camera motion); residual magnitude, Gaussian $\sigma{=}5$ (\texttt{motres\_extract.py}). \\
depth & MiDaS-small (torch hub, small transform), bicubic to $96{\times}56$; Depth-Anything-V2-Small (swept to weight 0). \\
DINO & \texttt{facebook/dinov2-with-registers-base}, input $448{\times}784$, last-layer CLS-to-patch attention averaged over the 12 heads, resized to $96{\times}56$, min-max normalised, rank; the rank map is blurred with $\sigma{=}3.5$ pixels and re-ranked before the exponent (\texttt{dino\_nextgen.py}). Local deviation (DEV): L2 norm of each last-layer patch token minus the frame-mean token, rank (\texttt{explore\_dinoagg\_frozen\_extract.py}). \\
defocus map $D$ & recipe estimator at $320{\times}320$, output resized to $96{\times}56$ (\texttt{v2\_finalize.py}). \\
\midrule \multicolumn{2}{l}{\emph{Fusion fit} (\texttt{stack\_v8\_fit.py}, \texttt{stack\_v9\_fit.py}, \texttt{sota\_plus\_final.py})} \\
objective and data & mean per-video sAUC on the Hollywood-2 train split only (823 videos, last frame of up to two clips each), twenty negative seeds. \\
search & coordinate ascent, one scalar at a time, repeated until no scalar moves: weights $\in\{0, 0.25, 0.5, 0.75, 1, 1.5, 2, 2.5, 3, 4, 5, 6, 8\}$; exponents $\in\{1, 1.5, 2, 3, 4, 6, 8, 12, 16, 24\}$; DINO blur $\sigma \in \{0, 2, 2.5, 3, 3.5, 4, 5, 6, 8\}$; DEV weight $\in\{0, 0.5, 1, 1.5, 2, 3, 4, 5, 6\}$. \\
supervised fusion & $r(M) + w^{*} r(S)$ with $w^{*} \in \{0.05, 0.1, 0.15, 0.2, 0.3, 0.4\}$ chosen on train per model, then frozen. \\
temporal variant & DINO, flow-residual and defocus maps replaced by the rank of their causal mean over lags $\{0,1,2,3,4,5,6,8,10,12,16\}$; scalars refitted with the same grids. \\
\midrule \multicolumn{2}{l}{\emph{Scoring} (\texttt{face\_gate.py}, \texttt{focusgate/metrics.py})} \\
sAUC & negatives drawn from the pooled fixation counts of all other videos in the set (the fixation distribution of other videos), twenty seeds, per-video mean, then the mean over videos. \\
intervals & paired per-video deltas, 10{,}000 percentile-bootstrap resamples over videos, seed 0; Holm correction over the sixteen Table-1 cells. \\
timing & one A100, batch 1, one forward pass per predicted frame, fp32 for the cues and the two 2019--2020 models, bf16 autocast for ViSAGE. \\
\bottomrule\end{tabular}\end{center}
\end{minipage}\par\medskip
\par\noindent\begin{minipage}{\textwidth}
\subsection*{I. Abstention under the two rank normalisations}
\begin{center}\includegraphics[width=0.6\textwidth]{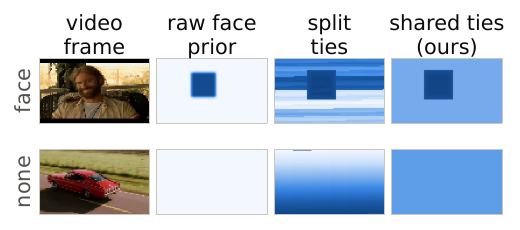}\end{center}
Fig.~S1. The face prior on two frames. Top: a detected face. Bottom: the detector returns nothing, so the raw prior is all zeros. Splitting ties fabricates a full-contrast pattern (standard deviation 0.29 across the map) that enters the sum at the cue's weight; sharing ties (midrank) gives the constant 0.5 and the fused term stays inert.
\end{minipage}\par\medskip
\twocolumn

%% file: refs.bib
@article{smith2012attentional,
  author={Smith, Tim J.},
  title={The attentional theory of cinematic continuity},
  journal={Projections},
  volume={6}, number={1}, pages={1--27}, year={2012}}

@article{zhang2019dof,
  author={Zhang, Tingting and Xia, Ling and Liu, Xiaofeng and Wu, Xiaoli},
  title={Eye movements during change detection: the role of depth of field},
  journal={Cognitive Computation and Systems},
  volume={1}, number={2}, pages={55--59}, year={2019}}

@article{vo2012eyes,
  author={V{\~o}, Melissa L.-H. and others},
  title={Do the eyes really have it? {D}ynamic allocation of attention when viewing moving faces},
  journal={Journal of Vision},
  volume={12}, number={13}, pages={3--3}, year={2012}}

@article{mital2011clustering,
  author={Mital, Parag K. and others},
  title={Clustering of gaze during dynamic scene viewing is predicted by motion},
  journal={Cognitive Computation},
  volume={3}, number={1}, pages={5--24}, year={2011}}

@inproceedings{shi2014discriminative,
  author={Shi, Jianping and others},
  title={Discriminative blur detection features},
  booktitle={Proc. IEEE CVPR},
  pages={2965--2972}, year={2014}}

@inproceedings{cun2020defocus,
  author={Cun, Xiaodong and Pun, Chi-Man},
  title={Defocus blur detection via depth distillation},
  booktitle={Proc. ECCV},
  pages={747--763}, year={2020}}

@article{kim2018defocus,
  author={Kim, Beomseok and others},
  title={Defocus and motion blur detection with deep contextual features},
  journal={Computer Graphics Forum},
  volume={37}, number={7}, pages={277--288}, year={2018}}

@article{bylinskii2018metrics,
  author={Bylinskii, Zoya and Judd, Tilke and Oliva, Aude and Torralba, Antonio and Durand, Fr{\'e}do},
  title={What do different evaluation metrics tell us about saliency models?},
  journal={IEEE Trans. PAMI},
  volume={41}, number={3}, pages={740--757}, year={2019}}

@article{mathe2015actions,
  author={Mathe, Stefan and Sminchisescu, Cristian},
  title={Actions in the eye: dynamic gaze datasets and learnt saliency models for visual recognition},
  journal={IEEE Trans. PAMI},
  volume={37}, number={7}, pages={1408--1424}, year={2015}}

@inproceedings{wang2018dhf1k,
  author={Wang, Wenguan and others},
  title={Revisiting video saliency: a large-scale benchmark and a new model},
  booktitle={Proc. IEEE CVPR},
  pages={4894--4903}, year={2018}}

@article{yunet,
  author={Wu, Wei and Peng, Hanyang and Yu, Shiqi},
  title={{YuNet}: a tiny millisecond-level face detector},
  journal = {Machine Intelligence Research},
  volume={20}, number={5}, pages={656--665}, year={2023}}

@inproceedings{hou2007saliency,
  author={Hou, Xiaodi and Zhang, Liqing},
  title={Saliency detection: a spectral residual approach},
  booktitle={Proc. IEEE CVPR},
  pages={1--8}, year={2007}}

@article{zhuo2011defocus,
  author={Zhuo, Shaojie and Sim, Terence},
  title={Defocus map estimation from a single image},
  journal={Pattern Recognition},
  volume={44}, number={9}, pages={1852--1858}, year={2011}}

@inproceedings{droste2020unisal,
  author={Droste, Richard and others},
  title={Unified image and video saliency modeling},
  booktitle={Proc. ECCV}, pages={419--435}, year={2020}}

@inproceedings{min2019tased,
  author={Min, Kyle and Corso, Jason J.},
  title={{TASED-Net}: temporally-aggregating spatial encoder-decoder network for video saliency detection},
  booktitle={Proc. IEEE ICCV}, pages={2394--2403}, year={2019}}

@article{ranftl2020midas,
  author={Ren{\'e} Ranftl and others},
  title={Towards robust monocular depth estimation: Mixing datasets for zero-shot cross-dataset transfer},
  journal = {IEEE Trans. PAMI},
  volume={44},
  number={3},
  year={2022}}

@inproceedings{liu2023evp,
  author={Weihuang Liu and Xi Shen and Chi-Man Pun and Xiaodong Cun},
  title={Explicit visual prompting for low-level structure segmentations},
  booktitle = {Proc. IEEE CVPR},
  year={2023}}

@inproceedings{jin2023ddffnet,
  author={Yuxin Jin and Ming Qian and Jincheng Xiong and Nan Xue and Gui-Song Xia},
  title={Depth and {DOF} cues make a better defocus blur detector},
  booktitle = {Proc. IEEE ICME},
  year={2023}}

@inproceedings{guo2008pqft,
  author={Chenlei Guo and Qi Ma and Liming Zhang},
  title={Spatio-temporal saliency detection using phase spectrum of quaternion {Fourier} transform},
  booktitle={Proc. IEEE CVPR},
  year={2008}}

@inproceedings{yang2024dav2,
  author={Lihe Yang and others},
  title={Depth {Anything} {V2}},
  booktitle = {Proc. NeurIPS},
  year={2024}}

@inproceedings{wang2026visage,
  author={Kun Wang and Yupeng Hu and Zhiran Li and Hao Liu and Qianlong Xiang and Liqiang Nie},
  title={{ViSAGE} @ {NTIRE} 2026 challenge on video saliency prediction},
  booktitle={Proc. IEEE/CVF CVPR Workshops},
  year={2026}}

@inproceedings{darcet2024registers,
  author = {Timoth{\'e}e Darcet and Maxime Oquab and Julien Mairal and Piotr Bojanowski},
  title = {Vision transformers need registers},
  booktitle = {Proc. ICLR},
  year = {2024}
}

@inproceedings{baveye2012blur,
  author={Yoann Baveye and Fabrice Urban and Christel Chamaret},
  title={Image and Video Saliency Models Improvement by Blur Identification},
  booktitle={Proc. ICCVG},
  year={2012}}

@article{oquab2023dinov2,
  author = {Maxime Oquab and Timoth{\'e}e Darcet and Th{\'e}o Moutakanni and others},
  title = {{DINOv2}: Learning Robust Visual Features without Supervision},
  journal = {Trans. Mach. Learn. Res.},
  year = {2024}
}

@inproceedings{chandra2025vinetpp,
  author = {Rohit Girmaji and Siddharth Jain and Bhav Beri and Sarthak Bansal and Vineet Gandhi},
  title = {Minimalistic Video Saliency Prediction via Efficient Decoder \& Spatio Temporal Action Cues},
  booktitle = {Proc. IEEE ICASSP},
  year = {2025}, pages = {1--5}
}

@article{alshawi2018uncertainty,
  author = {Tariq Alshawi and Zhiling Long and Ghassan AlRegib},
  title = {Unsupervised uncertainty estimation using spatiotemporal cues in video saliency detection},
  journal = {IEEE Trans. Image Process.},
  volume = {27}, number = {6}, pages = {2818--2827}, year = {2018}}

@article{cong2016depthconf,
  author = {Runmin Cong and Jianjun Lei and Changqing Zhang and Qingming Huang and Xiaochun Cao and Chunping Hou},
  title = {Saliency detection for stereoscopic images based on depth confidence analysis and multiple cues fusion},
  journal = {IEEE Signal Process. Lett.},
  volume = {23}, number = {6}, pages = {819--823}, year = {2016}}

@inproceedings{moradi2024salfom,
  author={Moradi, Morteza and Moradi, Mohammad and Rundo, Francesco and Spampinato, Concetto and Borji, Ali and Palazzo, Simone},
  title={{SalFoM}: dynamic saliency prediction with video foundation models},
  booktitle={Proc. ICPR}, year={2024}, pages={33--48}}

@inproceedings{xiong2024diffsal,
  author={Xiong, Junwen and Zhang, Peng and You, Tao and Li, Chuanyue and Huang, Wei and Zha, Yufei},
  title={{DiffSal}: joint audio and video learning for diffusion saliency prediction},
  booktitle={Proc. IEEE CVPR}, year={2024}}

@inproceedings{moskalenko2024aim,
  author={Moskalenko, Andrey and Bryncev, Alexey and Vatolin, Dmitry and others},
  title={{AIM} 2024 challenge on video saliency prediction: methods and results},
  booktitle={Proc. ECCV Workshops}, year={2024}}

@inproceedings{moskalenko2026ntire,
  author={Moskalenko, Andrey and Bryncev, Alexey and Kosmynin, Ivan and others},
  title={{NTIRE} 2026 challenge on video saliency prediction: methods and results},
  booktitle={Proc. IEEE/CVF CVPR Workshops}, year={2026}}
